\documentclass{article}
\usepackage{spconf,amsmath,graphicx}
\usepackage{booktabs}

\usepackage{enumitem}
\setlist{nosep, leftmargin=14pt}

\usepackage{mwe} 

\title{The Interplay of Demographic Composition and 
Supervision Budgets in Medical Imaging Foundation Models}

\name{\shortstack{Oscar Jimenez-del-Toro\textsuperscript{1} \qquad
Gokhan Ozbulak\textsuperscript{1} \qquad
Dilermando Queiroz\textsuperscript{1,2}\\[4pt]
Roberto Pulvirenti\textsuperscript{1} \qquad
Lilian Berton\textsuperscript{2} \qquad
Andre Anjos\textsuperscript{1}}}

\address{\textsuperscript{1}Idiap Research Institute, 
Martigny, Switzerland
\textsuperscript{2}Federal University of S\~ao Paulo, 
S\~ao Paulo, Brazil}
\begin{document}
%
\maketitle
\begin{abstract}
Medical imaging foundation models enable data-efficient adaptation to downstream tasks, but the combined effects of demographic training composition and supervision availability remain poorly understood. 
We investigate how these factors influence adaptation and cross-population generalization by fine-tuning specialist and generalist foundation models on chest X-ray and retinal image classification tasks. 
Using matched training sample counts and increasing positive-example budgets, we compare younger-only, older-only, and mixed-age training strategies.

We identify significant interactions between demographic training composition and positive supervision across all four model–task combinations. 
On chest X-rays, specialist and generalist models exhibit opposing shifts in the relative effectiveness of younger-only versus older-only training as supervision increases. 
On retinal images, both models show relative shifts toward older-only training. 
Comparisons with mixed-age fine-tuning further reveal supervision-dependent trade-offs in demographic data selection.

These findings demonstrate that demographic training advantages are not fixed but vary with supervision availability and model–task characteristics, highlighting the importance of jointly considering model choice, training population, and annotation budgets for reliable medical imaging adaptation.
\end{abstract}

\section{Introduction}
\label{sec:intro}
Medical imaging foundation models leverage large-scale pretraining to learn transferable representations for downstream clinical tasks \cite{engelmann2025retfoundgreen,yao2024evax}. 
Generalist models, such as DINOv3, learn broadly transferable representations, whereas specialist models, including EVA-X and RETFoundGreen, employ modality-specific pretraining \cite{simeoni2025dinov3,engelmann2025retfoundgreen,yao2024evax}. 
Although recent studies have compared their predictive performance and data efficiency \cite{zhou2025generalist,queiroz2024does}, reliable adaptation across heterogeneous datasets and patient populations remains challenging \cite{germani2026shortcut}.

Patient demographics introduce an additional dimension to foundation model adaptation. 
Age-related differences in disease prevalence and physiological characteristics may affect medical image appearance and model generalization \cite{hochhegger2012chest,orlov2019age}. 
While previous research has investigated demographic performance disparities, subgroup-aware learning, and utility--fairness trade-offs \cite{seyyed2021chexclusion,xu2026rethinking,pmlr-v267-alloula25a,ozbulak2025multiobjective}, whether the effectiveness of demographic-specific training changes with supervision availability remains largely unexplored. 
In particular, it is unclear whether these relationships differ between specialist and generalist foundation models or translate into practical advantages over mixed-demographic training.

To address this gap, we systematically investigate how demographic training composition and positive-example availability jointly influence foundation model adaptation and cross-population generalization. 
Across chest X-ray and retinal image classification tasks, we fine-tune specialist and generalist models using exclusively younger or older patients under matched training sample counts. 
By progressively increasing positive supervision, we examine demographic training-by-budget interactions and evaluate generalization across age groups. 
We further compare demographic-exclusive and mixed-age fine-tuning to assess the practical implications of demographic data selection. Both downstream datasets are distinct from the specialist models' pretraining sources.

Our contributions are threefold: (i) demonstrating that demographic training advantages vary with positive supervision; (ii) revealing distinct adaptation patterns across specialist and generalist foundation models, including cross-population generalization; and (iii) identifying supervision-dependent trade-offs between demographic-exclusive and mixed-age training.

Together, these findings demonstrate that the effectiveness of demographic training composition depends on supervision availability and model–task characteristics, informing data selection and population-aware evaluation in medical imaging. The source code will be made publicly available.
\section{Methods}
\label{sec:methods}
\subsection{Population-Dependent Adaptation and
Data-Selection Analysis}
Let $M_{a,i,n}$ denote foundation model $a$
fine-tuned on population $i$ with $n$ positive
examples, and $P(M_{a,i,n},g)$ its performance
on evaluation population $g$.
Training populations comprise younger-only
($Y$, $<60$ years), older-only ($O$, $\geq 60$ years),
and mixed-age ($X$) samples, using 60 years
as an operational age threshold.
Evaluation population $g$ denotes the overall,
younger, or older test population.

The demographic training advantage is defined as
\begin{equation}
D_{a,g}(n) =
P(M_{a,Y,n},g) - P(M_{a,O,n},g).
\end{equation}
Its dependence on positive supervision is measured
by the training-by-budget interaction
\begin{equation}
I_{a,g} =
D_{a,g}(n_{\max}) - D_{a,g}(n_{\min}).
\end{equation}
Positive values indicate a relative shift toward
younger-only training, and negative values toward
older-only training.
Specialist--generalist differences are quantified by
\begin{equation}
C_g =
I_{\mathrm{specialist},g} -
I_{\mathrm{generalist},g}.
\end{equation}
For specialist models, we compare demographic-exclusive and mixed-age training through
\begin{equation}
B_{a,i,g}(n) =
P(M_{a,i,n},g) - P(M_{a,X,n},g),
\quad i \in \{Y,O\}.
\end{equation}
Positive values favor exclusive training, whereas
negative values favor mixed-age training.
The corresponding budget interaction is
\begin{equation}
J_{a,i,g} =
B_{a,i,g}(n_{\max}) -
B_{a,i,g}(n_{\min}).
\end{equation}
Interactions use the smallest and largest budgets
shared by the compared configurations.
\subsection{Statistical Analysis}
Uncertainty is estimated using 20,000 paired
patient-level bootstrap resamples of the shared
test set, stratified by age and patient-level
disease status (at least one positive image).
Each resample retains all images from selected
patients and is applied identically across model
predictions to preserve pairing.
Confidence intervals (CIs) are constructed
from bootstrap percentiles.

For overall average precision (AP), Bonferroni
correction controls the family-wise error rate
at 0.05 separately for two four-comparison
interaction families (younger-versus-older and
exclusive-versus-mixed training; 98.75\% CIs)
and twelve exclusive-versus-mixed fixed-budget
contrasts (99.58\% CIs).

Specialist--generalist interaction contrasts,
age-specific analyses, demographic performance
gaps, and AUROC analyses are exploratory, using
nominal 95\% CIs.
AP is the primary metric, with AUROC as a
secondary metric.
\section{Experimental Setup}
\label{sec:experiments}
We evaluate two medical imaging classification tasks,
comparing specialist and generalist foundation models
under demographic-specific fine-tuning, with mixed-age
training additionally assessed for specialist models.
\subsection{PadChest Cardiomegaly Classification}
PadChest comprises 89,890 posterior-anterior chest
X-ray studies after excluding samples with missing
data \cite{bustos2020padchest}.
For binary cardiomegaly classification, we compare
EVA-X-B, a chest radiography foundation model,
with DINOv3-B.
Positive prevalence is substantially higher in
older patients (6.2\%) than younger patients (0.9\%).
\subsection{mBRSET Diabetic Retinopathy Classification}
mBRSET comprises 4,147 color fundus photographs
with clinical metadata after excluding samples
with missing data \cite{boeck2023mbrset}.
For binary diabetic retinopathy classification
based on International Clinical Diabetic
Retinopathy (ICDR) grading, we compare
RETFound-Green \cite{engelmann2025retfoundgreen}
with DINOv3-B.
Positive prevalence is more comparable between
younger (11.0\%) and older (8.7\%) patients.

Both downstream datasets are distinct from
the specialist models' pretraining datasets,
enabling evaluation of demographic adaptation
under dataset transfer.
\subsection{Fine-tuning and Evaluation Protocol}
We use age as a clinically motivated proof of concept for demographic composition, reflecting differences in disease prevalence and age-related imaging characteristics. 
Patients are divided into younger
($<60$ years) and older ($\geq 60$ years) groups,
consistent with established population-aging
thresholds.

All models are independently fully fine-tuned
on younger-only or older-only samples, while
specialist models are additionally fine-tuned
on mixed-age samples as a practical
data-selection baseline.

The age group with fewer available samples
determines the maximum training count for each
class. Positive and negative counts are matched
across configurations at shared budgets.
Negative counts remain fixed at 5,000 for PadChest and 1,000 for mBRSET, while positive-example budgets range from 100 to 500 and 100 to 400,
respectively.
Comparisons are restricted to shared budget levels.

Both datasets use patient-level 70\%/15\%/15\%
training/validation/test splits.
Validation and test sets remain fixed across
configurations, preserving both age groups
at their natural disease prevalence.
Checkpoints are selected by minimum overall
validation loss.
%
\begin{figure*}[t]
\centering
\includegraphics[width=0.75\textwidth]{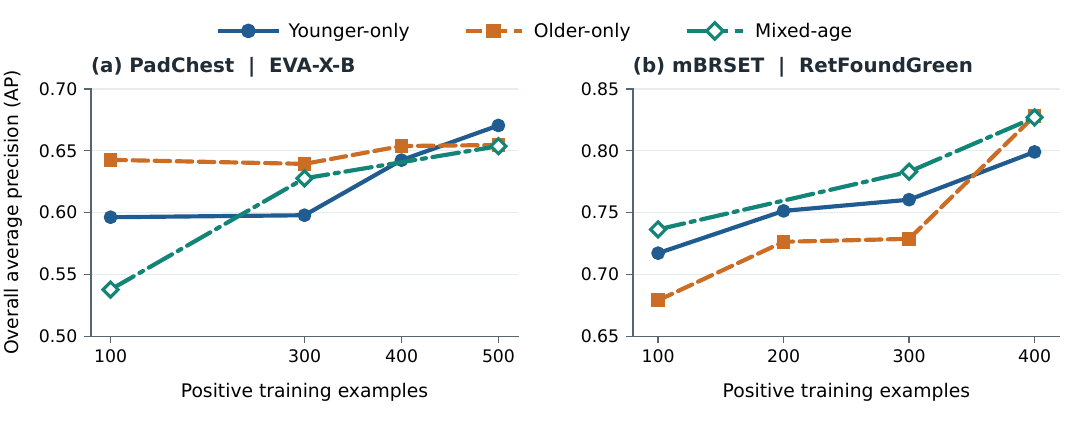}
\caption{\textbf{Overall adaptation performance under
demographic training composition.}
Test-set average precision (AP) of specialist models:
(a) EVA-X-B on PadChest and (b) RETFound-Green on mBRSET,
following younger-only, older-only, and mixed-age
fine-tuning across increasing positive-example budgets.
Training sample counts are matched across configurations
at shared budgets, with fixed negative counts.
All models are evaluated on the same held-out test set
within each task. 
Markers indicate evaluated budgets; lines guide the eye. 
Statistical comparisons with mixed-age training are reported in the Results.}
\label{fig:adaptation_performance}
\end{figure*}
\section{Results}
\label{sec:results}
\subsection{Adaptation under Increasing Positive Supervision}
Overall AP trajectories revealed differences in adaptation across models and training configurations
(Fig.~\ref{fig:adaptation_performance}).
On PadChest, EVA-X-B consistently outperformed DINOv3-B. 
Younger-only and mixed-age EVA-X-B
improved from 0.596 to 0.670 and 0.538 to 0.654, respectively, between 100 and 500 positives.
In contrast, younger-only DINOv3-B peaked at 300 positives (AP = 0.426) before declining.

On mBRSET, RETFound-Green improved with increasing supervision across all training configurations.
Mixed-age and older-only AP increased from 0.736 to 0.827 and 0.679 to 0.828, respectively, between 100 and 400 positives.
Both demographic-specific DINOv3-B configurations instead peaked at 300 positives.
Mixed-age AUROC also increased for EVA-X-B (0.927 to 0.939) and RETFoundGreen
(0.861 to 0.917), consistent with their AP trends.
\subsection{Population-Dependent Adaptation and Generalization}
All four overall AP demographic training-by-budget
interactions remained significant after Bonferroni
correction (98.75\% CIs;
Table~\ref{tab:budget_interactions}, Panel A).
These interactions quantify how the relative advantage
of younger-only versus older-only training changes
with positive supervision.

On PadChest, EVA-X-B exhibited a positive interaction
($+0.062$, adjusted CI $[0.039, 0.085]$),
indicating a shift from an older-only to a
younger-only training advantage.
Conversely, DINOv3-B exhibited a negative interaction
($-0.077$, $[-0.101, -0.051]$), reflecting a
relative shift toward older-only training.
Their specialist--generalist interaction contrast
was $+0.140$ (nominal 95\% CI $[0.113, 0.164]$; Panel C), supporting distinct adaptation patterns.
AUROC showed opposing interaction directions ($+0.0067$ versus $-0.0186$), with a positive
model contrast ($+0.025$, nominal 95\%
CI $[0.018, 0.033]$).

On mBRSET, both RETFoundGreen and DINOv3-B exhibited negative AP interactions (approximately
$-0.067$), indicating relative shifts toward older-only training. 
RETFoundGreen consequently
reversed its initial younger-only advantage.
Their direct interaction contrast was negligible ($+0.0002$, nominal 95\% CI $[-0.057, 0.051]$), providing no evidence of different adaptation
interactions. 
AUROC interactions were similarly
negative ($-0.026$ and $-0.028$), without an established specialist--generalist difference.

AP and AUROC interaction directions agreed across all four model--task combinations.
Exploratory age-stratified analyses further revealed differences in cross-population generalization, with specialist--generalist AP interaction contrasts exhibiting nominal
95\% CIs excluding zero in both evaluation groups on PadChest, but not on mBRSET.
\subsection{Demographic-Exclusive vs. Mixed-Age Fine-tuning}
We compared demographic-exclusive and mixed-age
fine-tuning under matched positive-example budgets
(Fig.~\ref{fig:adaptation_performance}).
On PadChest, younger-only and older-only EVA-X-B
significantly outperformed mixed-age training
at 100 positives ($\Delta\mathrm{AP}=+0.058$
and $+0.105$, respectively), with AUROC differences
in the same direction.
At 300 positives, mixed-age training outperformed
younger-only training ($\Delta\mathrm{AP}=-0.030$),
although the AUROC difference was inconclusive.
Neither exclusive configuration showed a significant
AP advantage at 500 positives.

On mBRSET, mixed-age RETFound-Green significantly
outperformed older-only training at 300 positives
($\Delta\mathrm{AP}=-0.054$), supported by
AUROC ($-0.038$, nominal 95\% CI excluding zero).
No significant AP differences were found at
100 or 400 positives. Overall, four of twelve
fixed-budget AP contrasts survived Bonferroni
correction (99.58\% CIs).

Exclusive-versus-mixed AP budget interactions
(Table~\ref{tab:budget_interactions}, Panel B)
revealed decreasing younger-only and older-only
advantages on PadChest ($-0.042$ and $-0.104$,
respectively). Conversely, older-only RETFound-Green
improved relative to mixed-age training on mBRSET
($+0.059$), while the younger-only interaction
was not significant. All three interactions
remained significant after Bonferroni correction
(98.75\% CIs).

AUROC supported the opposing older-only versus
mixed-age interaction directions on PadChest
($-0.0085$) and mBR- SET ($+0.0283$), with nominal 95\% CIs excluding zero.
These findings establish supervision-dependent trade-offs between demographic-exclusive and mixed-age fine-tuning.
%
%
\begin{table}[t]
\centering
\caption{Overall AP training-by-budget interactions:
(A) younger-only versus older-only training,
(B) demographic-exclusive versus mixed-age training,
and (C) specialist--generalist contrasts.
Positive values indicate a relative shift toward
younger-only training in (A), exclusive training in (B), or the specialist model in (C).
Budgets range from 100--500 positives on PadChest and 100--400 on mBRSET.
CIs are Bonferroni-adjusted 98.75\% in (A,B) and nominal 95\% in (C).
Y, O, and X denote younger-only, older-only, and mixed-age training, respectively.}
\label{tab:budget_interactions}
\setlength{\tabcolsep}{3pt}
\footnotesize
\begin{tabular}{@{}llrl@{}}
\hline
Dataset & Comparison & $\Delta\mathrm{AP}$ & Confidence interval \\
\hline
\multicolumn{4}{@{}l}{\textit{A. Younger-only minus older-only}} \\
PadChest & EVA-X-B
         & $+0.062$ & $[+0.039,+0.085]$ \\
         & DINOv3-B
         & $-0.077$ & $[-0.101,-0.051]$ \\
mBRSET   & RetFoundGreen
         & $-0.067$ & $[-0.132,-0.012]$ \\
         & DINOv3-B
         & $-0.067$ & $[-0.113,-0.019]$ \\
\hline
\multicolumn{4}{@{}l}{\textit{B. Exclusive minus mixed-age}} \\
PadChest & Y--X (EVA-X-B)
         & $-0.042$ & $[-0.079,-0.002]$ \\
         & O--X (EVA-X-B)
         & $-0.104$ & $[-0.133,-0.071]$ \\
mBRSET   & Y--X (RetFoundGreen)
         & $-0.009$ & $[-0.071,+0.050]$ \\
         & O--X (RetFoundGreen)
         & $+0.059$ & $[+0.007,+0.113]$ \\
\hline
\multicolumn{4}{@{}l}{\textit{C. Specialist minus generalist interaction}} \\
PadChest & Specialist--DINOv3-B
         & $+0.140$ & $[+0.113,+0.164]$ \\
mBRSET   & Specialist--DINOv3-B
         & $+0.0002$ & $[-0.057,+0.051]$ \\
\hline
\end{tabular}
\end{table}
%
%
%
%
%
\section{Discussion and Implications}
\label{sec:discussion}
Demographic training composition interacted significantly with positive supervision across all four model--task combinations, although the direction of these interactions varied. 
On PadChest, EVA-X-B and DINOv3-B exhibited opposing shifts in relative demographic training effectiveness, supported by an exploratory between-model contrast. 
On mBRSET, both models shifted toward older-only training, without an established between-model difference. 
AUROC interactions followed the same directions as AP. 
Together, these results indicate that demographic training advantages evolve with supervision availability in a model--task-dependent manner.

The datasets exhibited contrasting age-related disease prevalence patterns, with cardiomegaly more prevalent among older patients and diabetic retinopathy more evenly distributed across age groups. 
Although training class counts were matched between demographic configurations, differences in disease presentation and imaging characteristics may contribute to the observed adaptation patterns. 
Both specialist models were also evaluated on downstream datasets distinct from their pretraining sources, demonstrating the relevance of demographic training composition in transfer settings without establishing its underlying mechanisms.

Comparisons with mixed-age fine-tuning revealed additional supervision-dependent trade-offs. 
On PadChest, demographic-exclusive EVA-X-B outperformed mixed-age training at the smallest positive budget, but this advantage diminished with increasing supervision. 
On mBRSET, mixed-age RETFoundGreen outperformed older-only training at an intermediate budget, whereas older-only training improved relative to mixed-age training as supervision increased. 
These results suggest that the preferred demographic training composition may depend on annotation availability rather than following a fixed data-selection strategy.

From a practical perspective, demographic composition and positive supervision should be considered jointly when selecting fine-tuning data. 
Subgroup-specific training may offer advantages under limited supervision that diminish or reverse with additional positive examples. Furthermore, improvements in aggregate performance may not translate uniformly across evaluation age groups, underscoring the importance of assessing cross-population generalization alongside overall performance.

These findings complement previous research on foundation model transferability and subgroup-aware learning \cite{queiroz2024does,pmlr-v267-alloula25a}. 
Differences in pretrained representations and population-specific image characteristics may contribute to the observed patterns, although their mechanisms remain unexplored.

Several limitations warrant consideration. 
Increasing positive-example budgets while holding negative counts fixed changes both training-set size and class prevalence, while demographic composition varies jointly across positive and negative samples. 
Bootstrap inference captures test-set uncertainty but not training variability. 
Experiments were limited to two imaging tasks and age-defined populations, and observed model differences cannot be attributed solely to pretraining or architecture. 
Future work should investigate representation-level mechanisms, training variability, and alternative demographic mixtures across additional models, tasks, and external cohorts.
\section{Conclusion}
This study demonstrates that the effectiveness of demographic-specific fine-tuning varies with positive supervision and model--task characteristics. 
Significant demographic training-by-budget interactions and comparisons with mixed-age training reveal supervision-dependent differences in adaptation and data-selection effectiveness. 
These findings highlight the importance of jointly considering foundation model choice, demographic training composition, annotation budgets, and cross-population generalization in data-efficient medical imaging adaptation.
\section*{Compliance with Ethical Standards}
This retrospective study used de-identified
PadChest and mBRSET data, originally collected
under institutional ethical approval.
PadChest was publicly accessed, while mBRSET
was accessed through the required credentialed
procedure under its data-use conditions.
No new patient data were collected or
participants recruited.
\section*{Acknowledgments}
OpenAI's ChatGPT assisted with writing, restructuring,
and clarity improvements throughout the manuscript.
The authors reviewed all AI-assisted content and
retain full responsibility for its accuracy.
%
%
%
%
\bibliographystyle{IEEEbib}
\bibliography{strings,refs}
\end{document}